\def\year{2020}\relax
\documentclass[letterpaper]{article} 
\usepackage{aaai20}  
\usepackage{times}  
\usepackage{helvet} 
\usepackage{courier}  
\usepackage[hyphens]{url}  
\usepackage{graphicx} 
\usepackage{graphicx}  
\usepackage[ruled,linesnumbered]{algorithm2e}
\usepackage{makecell}
\usepackage{multirow}
\title{DARB: A Density-Adaptive Regular-Block Pruning for Deep Neural Networks}
\author{
Ao Ren\textsuperscript{$\dagger$$\ddagger$}\thanks{The work was done during Ao's internship at Alibaba DAMO Academy.},
Tao Zhang\textsuperscript{$\dagger$},
Yuhao Wang\textsuperscript{$\dagger$},
Sheng Lin\textsuperscript{$\ddagger$},
Peiyan Dong\textsuperscript{$\ddagger$}, \\
\bf \Large Yen-kuang Chen\textsuperscript{$\dagger$},
Yuan Xie\textsuperscript{$\dagger$},
Yanzhi Wang\textsuperscript{$\ddagger$} \\\\
\textsuperscript{$\dagger$}Alibaba DAMO Academy\\
\textsuperscript{$\ddagger$}Northeastern University\\
ren.ao@husky.neu.edu, \{t.zhang, yuhao.w\}@alibaba-inc.com, \{lin.sheng, dong.pe\}@husky.neu.edu \\
\{yk.chen, y.xie\}@alibaba-inc.com, yanz.wang@northeastern.edu
}

\newcommand{\citet}[1]{\citeauthor{#1}~\shortcite{#1}}
\newcommand{\citep}{\cite}

\newcommand{\darb}{DARB}  
\newcommand{\bmwm}{BMWM}  

\begin{document}

\maketitle

\begin{abstract}

The rapidly growing parameter volume of deep neural networks (DNNs) hinders the artificial
intelligence applications on resource constrained devices, such as mobile and wearable devices.
Neural network pruning, as one of the mainstream model compression techniques, is under extensive
study to reduce the model size and thus the amount of computation. And thereby, the state-of-the-art
DNNs are able to be deployed on those devices with high runtime energy efficiency. In contrast to
irregular pruning that incurs high index storage and decoding overhead, structured pruning
techniques have been proposed as the promising solutions. However, prior studies on structured
pruning tackle the problem mainly from the perspective of facilitating hardware implementation,
without diving into the deep to analyze the characteristics of sparse neural networks. The neglect
on the study of sparse neural networks causes inefficient trade-off between regularity and pruning
ratio. Consequently, the potential of structurally pruning neural networks is not sufficiently
mined.

In this work, we examine the structural characteristics of the irregularly pruned weight
matrices, such as the diverse redundancy of different rows, the sensitivity of different rows to
pruning, and the position characteristics of retained weights. By leveraging the gained insights as
a guidance, 
we first propose the novel {\em block-max weight masking (\bmwm)} method, which can effectively retain the salient 
weights while imposing high regularity to the weight matrix.  
As a further optimization, we propose a {\em density-adaptive regular-block (\darb)} pruning that can effectively take advantage of
the intrinsic characteristics of neural networks, and thereby outperform prior structured pruning
work with high pruning ratio and decoding efficiency. Our experimental results show that \darb~can
achieve 13$\times$ to 25$\times$ pruning ratio, which are 2.8$\times$ to 4.3$\times$ improvements
than the state-of-the-art counterparts on multiple neural network models and tasks. Moreover,
\darb~can achieve 14.3$\times$ decoding efficiency than block pruning with higher pruning
ratio.

\end{abstract}

\section{Introduction}

With the rapid development of deep neural networks (DNNs), artificial intelligence (AI) has
penetrated into various application domains. The success of the state-of-the-art deep learning
models highly relies on the large number of parameters and the resulting extensive computations. And
the trend of new models is to grow even larger and deeper. However, the enormous number of
parameters and computations prevents the models from being widely deployed in the scenarios that
have limited hardware resources, such as running these models on mobile or edge devices. To address
this issue, a lot of research efforts have been paid to reduce the model size and accelerate the
inference process of DNNs
\cite{Han2016a,park2017weighted,zhou2017incremental,leng2017extremely,wang2018towards,ren2019admm,ye2019adversarial}.

Neural network pruning is one of the major compression techniques to reduce model size by
removing the least important weights, i.e., the weights with small absolute value. 
It has been empirically proved in prior works \cite{Han2016a,ren2019admm} that such
irregular pruning can achieve the highest compression ratio than all other compression techniques
for neural networks, such as structured pruning, feature/activation pruning, and weight
quantization. Nonetheless, irregular pruning is mostly opted out in reality due to its notorious
positional irregularity of retained weights, which incurs inefficient index decoding and high index storage overhead.
Structured pruning techniques are developed to address the drawbacks of
irregular pruning. They reduce indexing overhead and achieve acceleration mainly through the
trade-off among regularity, pruning ratio, and model accuracy. That is, given the same accuracy,
the looser constraints on the regularity of the weights, the higher pruning ratio can be achieved.
However, little attention has been given to the intrinsic characteristics of neural networks for efficient trade-off. 

In this work, we endeavor to find a more effective structured pruning method by studying the
structural characteristics of the irregularly pruned weight matrix. In particular, we investigate the
distribution of \emph{row density} in the matrix and its correlation with the pruning
ratio\footnote{Row density is defined as the percentage of retained weights in a row after pruning, which
is the compliment of sparsity.}. Our study focuses on recurrent neural networks (RNNs), such as long
short term memory (LSTM) and gated recurrent unit (GRU), and the fully-connected layers of 
convolutional neural networks (CNNs). Our study reveals that: (i) maintaining the variety of row
density helps sustain model accuracy; (ii) the dense rows are more sensitive than the sparse rows to
further pruning; 
(iii) when dividing all rows into equally sized blocks and selecting in each block one weight with the largest magnitude, 
these locally salient weights have similar salience to the weights globally selected over the whole weight matrix.
Based on the gained insights, we innovatively propose a {\em block-max weight masking (\bmwm)} method 
and a {\em density-adaptive regular-block (\darb)} pruning method
that can achieve high pruning ratio, low index storing cost, efficient index decoding, and sustained
accuracy, \emph{simultaneously}. Our experimental results show that the proposed \darb~significantly
outperforms the state-of-the-art counterparts by up to 4.3$\times$ higher pruning ratio on various neural network
models and tasks. Moreover, \darb~can beat block pruning by 14.3$\times$ higher index decoding
efficiency and meanwhile achieve higher pruning ratio.

\begin{figure}[tbp]
\centering
\includegraphics[width=0.41\textwidth]{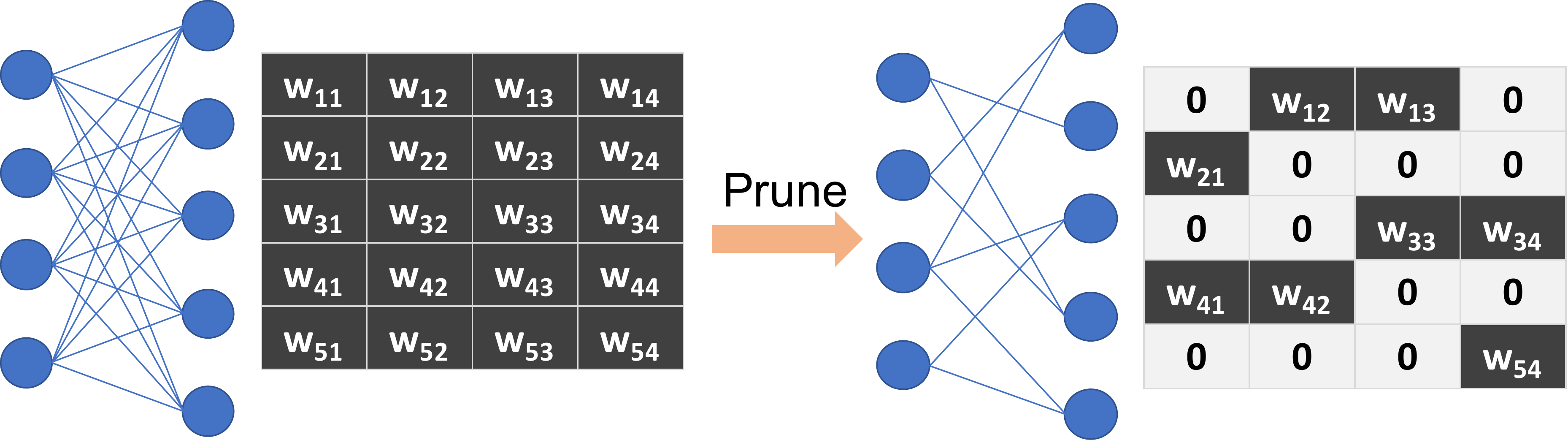}
\caption{Neural network pruning.} \label{fig:pruning}
\end{figure}


In summary, the main contributions of this work are:
\begin{itemize}
  \item We analyze the irregularly pruned weight matrices and 
  figure out that each row intrinsically has different density and denser rows 
  are more sensitive to further pruning, which requires
  attention to be paid when developing pruning algorithms.
  \item We study the positional characteristic of weights, and propose the 
  block-max weight masking (\bmwm) method that is more effective in retaining the salient weights
  than prior structured pruning methods.
  \item We propose the density-adaptive regular-block (DARB) pruning to achieve high regularity,
  high pruning ratio, and sustained accuracy, simultaneously.
\end{itemize}

\section{Background and Related Work}

Neural networks are becoming larger and deeper to achieve the state-of-the-art performance in many
domains, such as \cite{krizhevsky2012imagenet,simonyan2014very,he2016identity,szegedy2017inception}
in computer vision (CV) and \cite{vaswani2017attention,devlin2018bert,yang2019xlnet} in natural
language processing (NLP). Nonetheless, the large number of parameters and the resulting
computations impede the deployment of the models on devices that have limited on-chip resources.
Consequently, neural network compression techniques have gained increasing attraction in both
industry and academia. There are two major compression techniques, \emph{pruning} and
\emph{quantization}, where the former is mainly intended to reduce the redundancy existing in the
number of weights \cite{Han2016a,frankle2018lottery,ren2019admm} while the latter is to reduce the
redundancy in the representation precision of each single weight
\cite{leng2017extremely,park2017weighted,zhou2017incremental}. These two techniques are orthogonal
to each other and often combined to achieve the best compression ratio. This work only concentrates
on the pruning technique, which solely can achieve higher compression ratio.

\subsubsection{Irregular Pruning and Index Decoding}

\noindent As illustrated in Figure~\ref{fig:pruning}, pruning refers to removing the connections
between the neurons of two adjacent layers, which results in a sparse weight matrix. Since the
removed weights do not need to be stored and involved in computation, pruning is able to reduce
storage and computation overhead, which is critical for resource constrained devices.

\begin{figure}[tbp]
\centering
\includegraphics[width=0.43\textwidth]{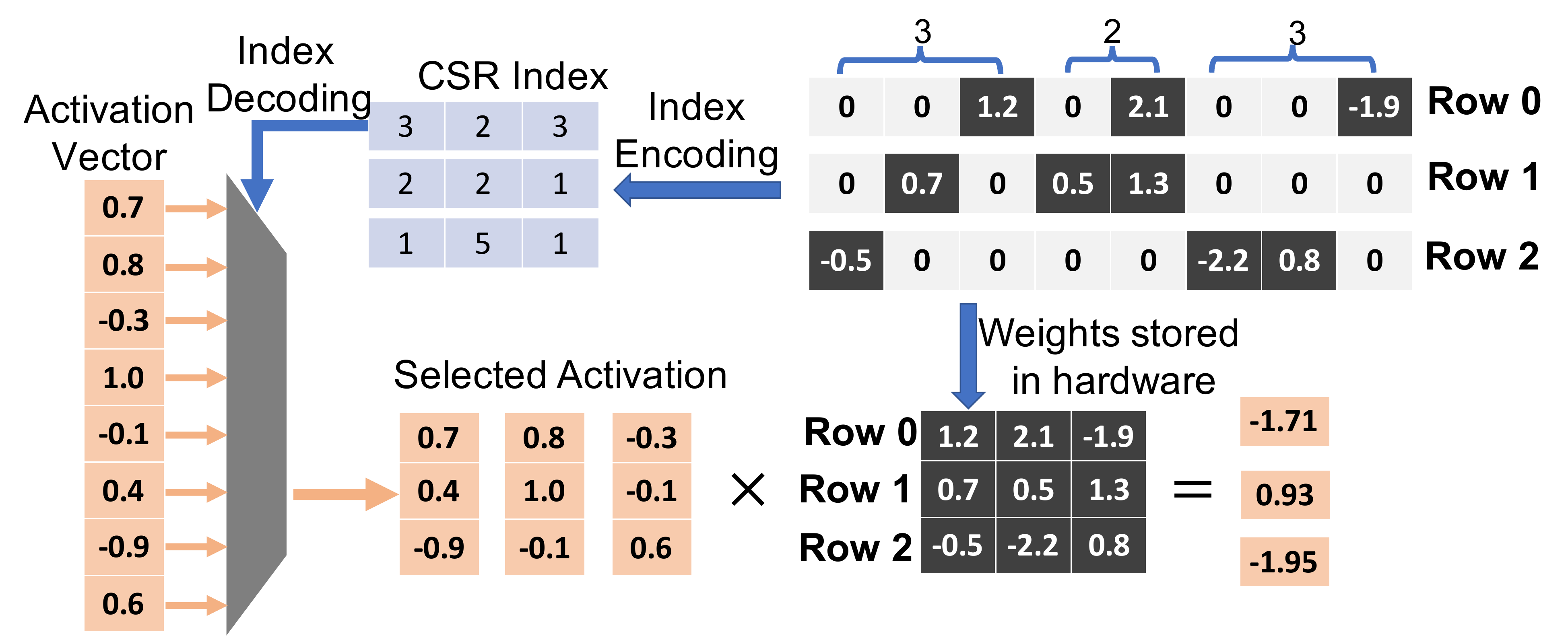}
\caption{Index encoding and decoding process.} \label{fig:encode_decode}
\end{figure}

Irregular pruning is the most straightforward pruning technique. The idea behind is that
the important weights have larger magnitude (i.e., absolute value),
so keeping only the top-K weights in magnitude should have little impact on accuracy. 
Although it can achieve impressive high pruning ratio
\cite{Han2016a,frankle2018lottery,ren2019admm}, the positions of the pruned weights are rather
random, and consequently, a large number of indices are required to record the positions of the retained weights. 
To save the index storage, the relative compressed sparse row (CSR) format \cite{Han2016a} is usually adopted,
which encodes each index by the relative distance (i.e., the number of zeros) between two
adjacent non-zero weights. Besides, a decoding process is needed to select the corresponding
activations for the retained weights. Figure~\ref{fig:encode_decode} illustrates
the encoding and decoding process. The main drawback of irregular pruning is that decoding one index
requires a search over the whole activation vector, and thus it brings little
acceleration and even speed degradation.

\subsubsection{Structured Pruning and Related Work}

\begin{figure*}[t]
\centering
\minipage{0.245\textwidth}
  \includegraphics[width=\linewidth]{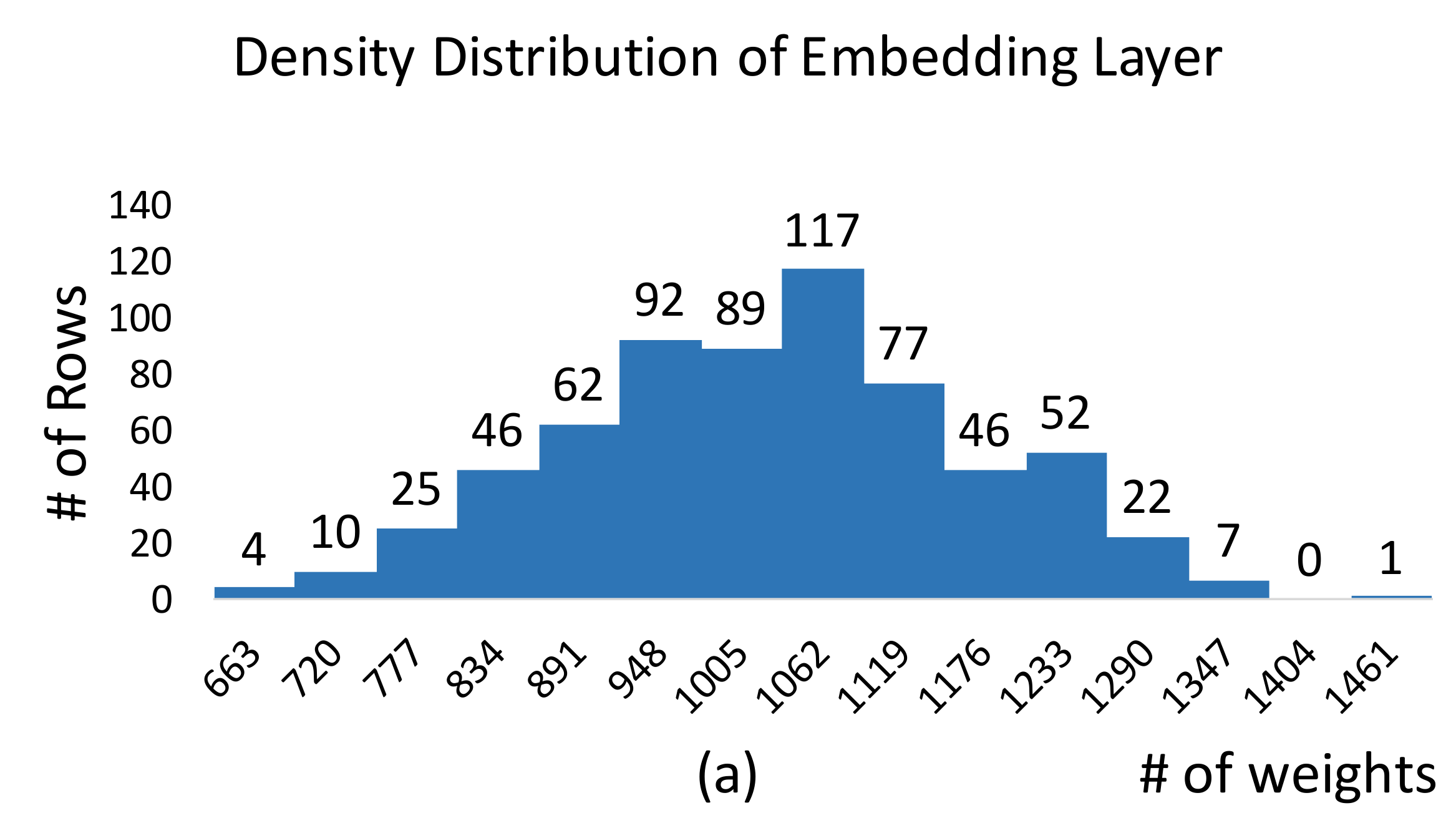}
\endminipage
\minipage{0.245\textwidth}
  \includegraphics[width=\linewidth]{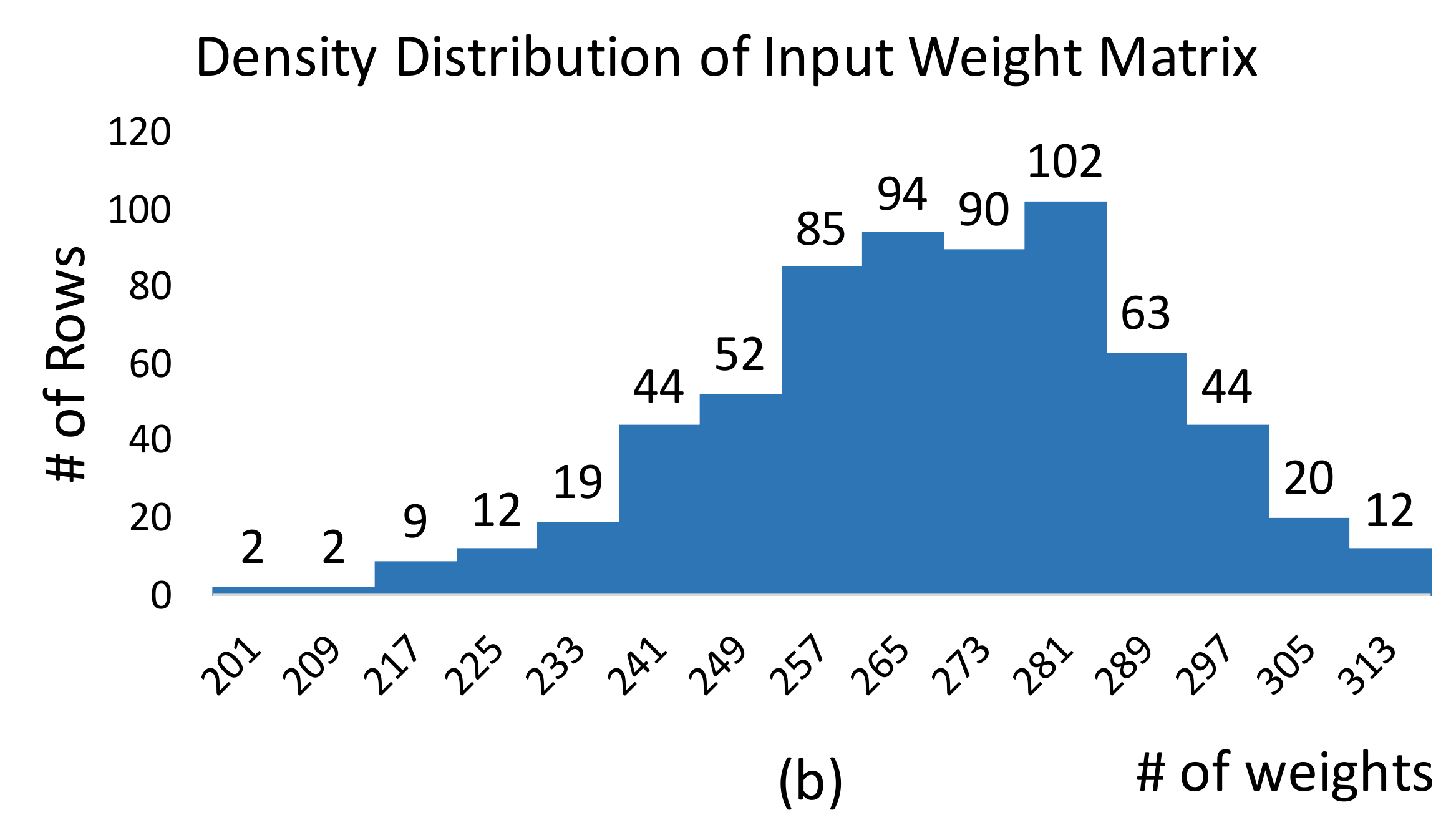}
\endminipage
\minipage{0.245\textwidth}%
  \includegraphics[width=\linewidth]{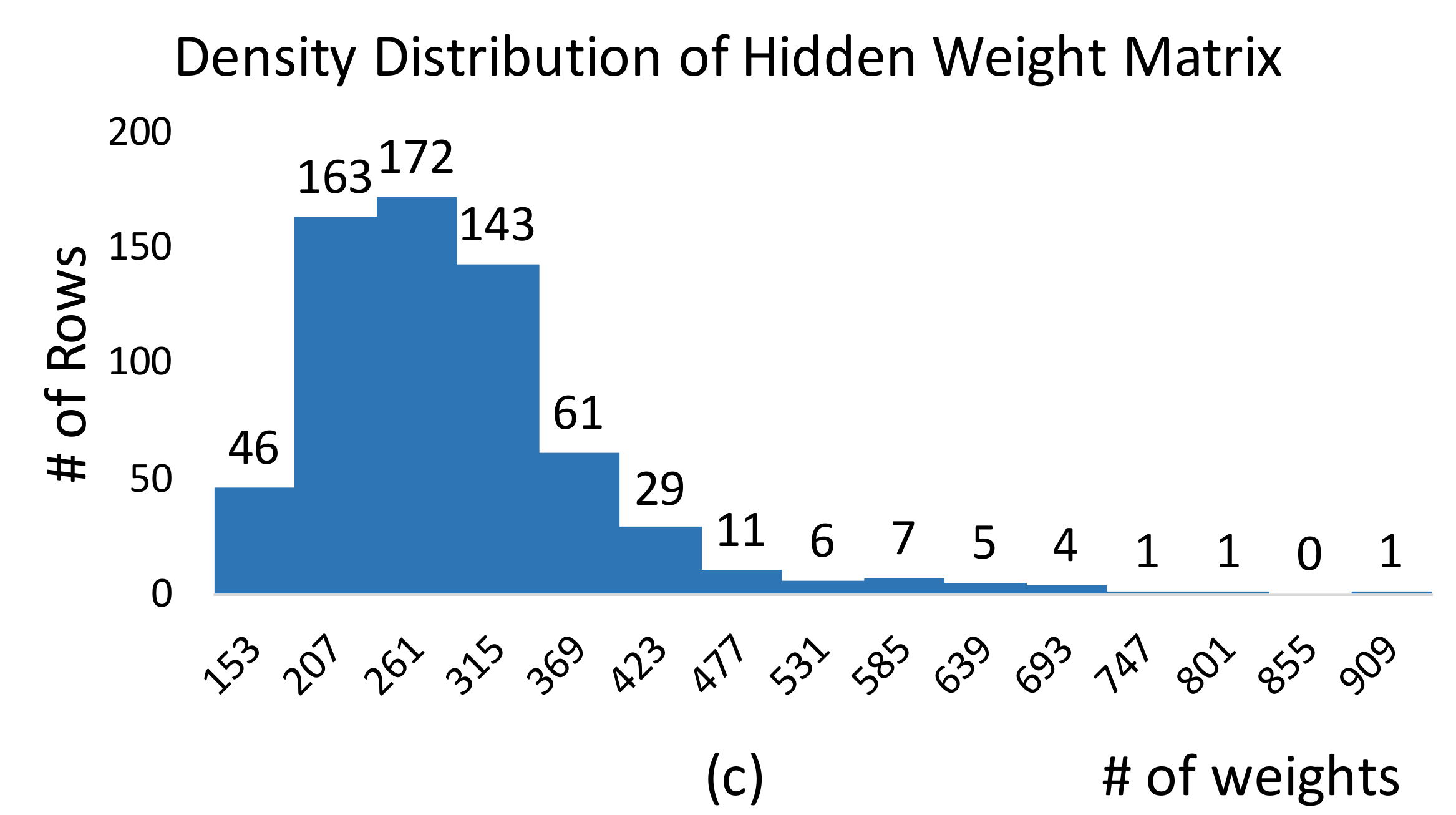}
\endminipage
\minipage{0.245\textwidth}%
  \includegraphics[width=\linewidth]{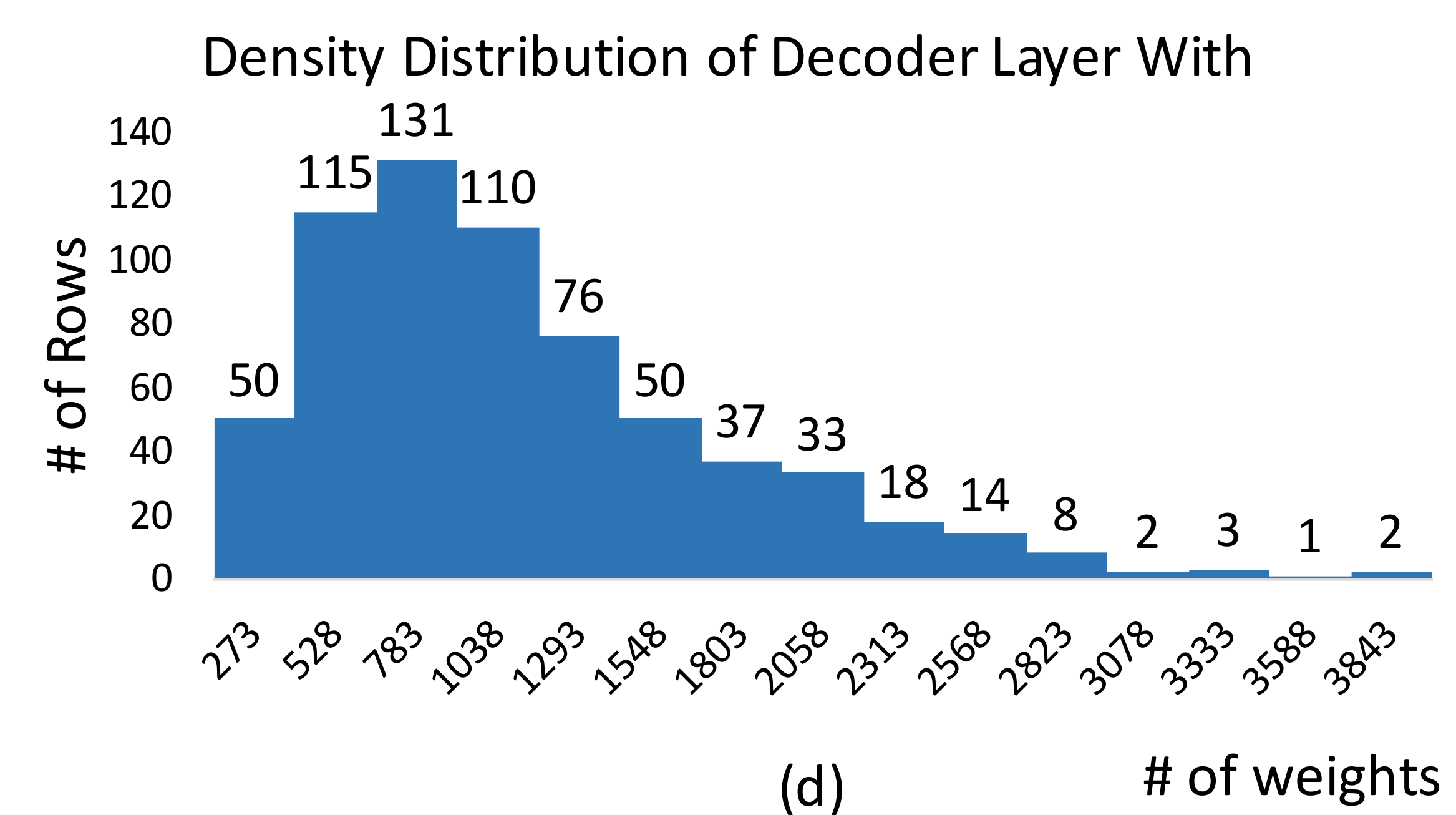}
\endminipage
\caption{Row density distribution of the four weight components of a medium LSTM with 90\% sparsity: (a) embedding layer, 
(b) input weight matrices, (c) hidden weight matrices, (d) decoder layer.}\label{fig:lstm_90}
\end{figure*}

\begin{figure*}[t]
\centering
\minipage{0.313\textwidth}
  \includegraphics[width=\linewidth]{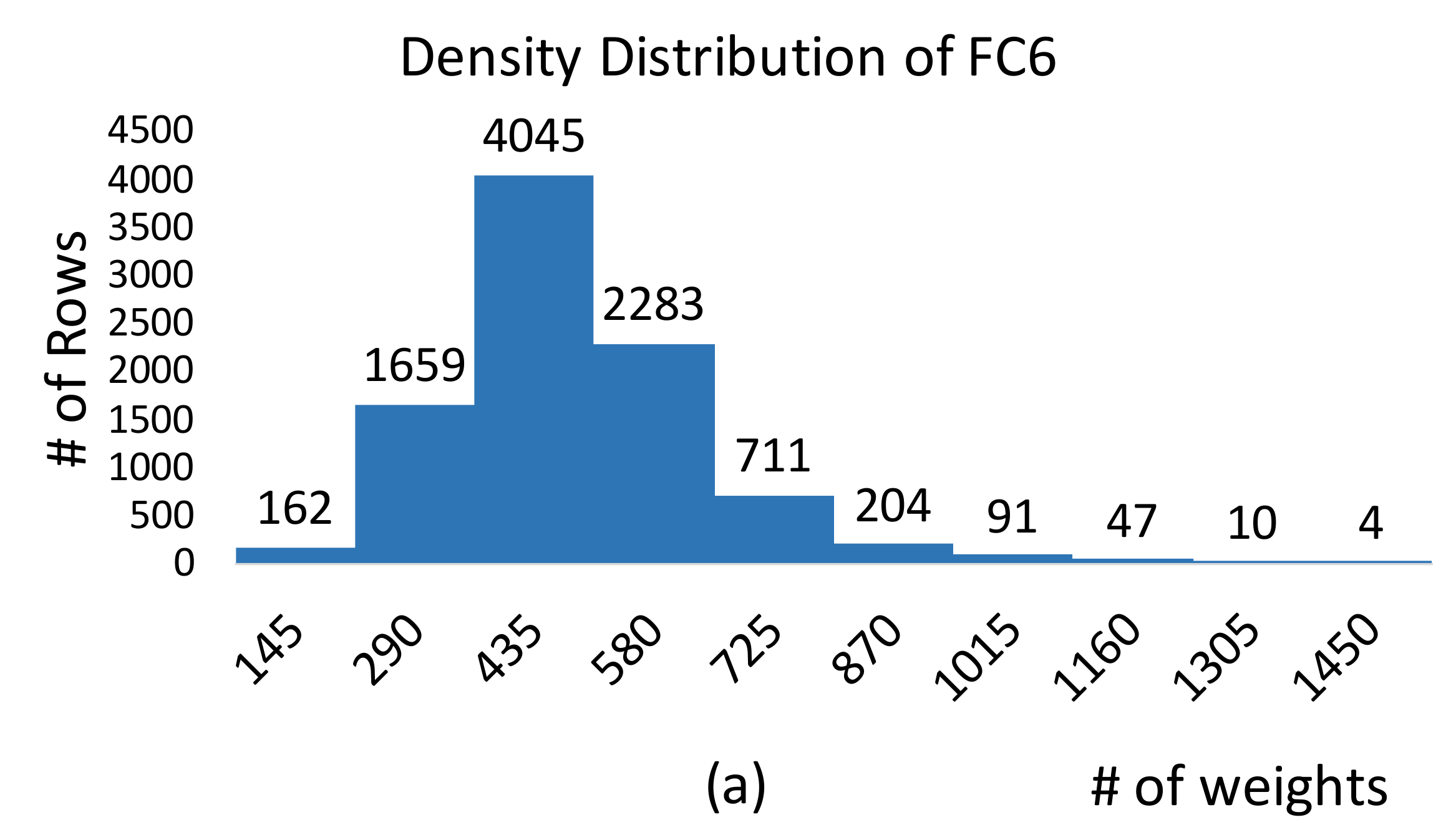}
\endminipage
\minipage{0.313\textwidth}
  \includegraphics[width=\linewidth]{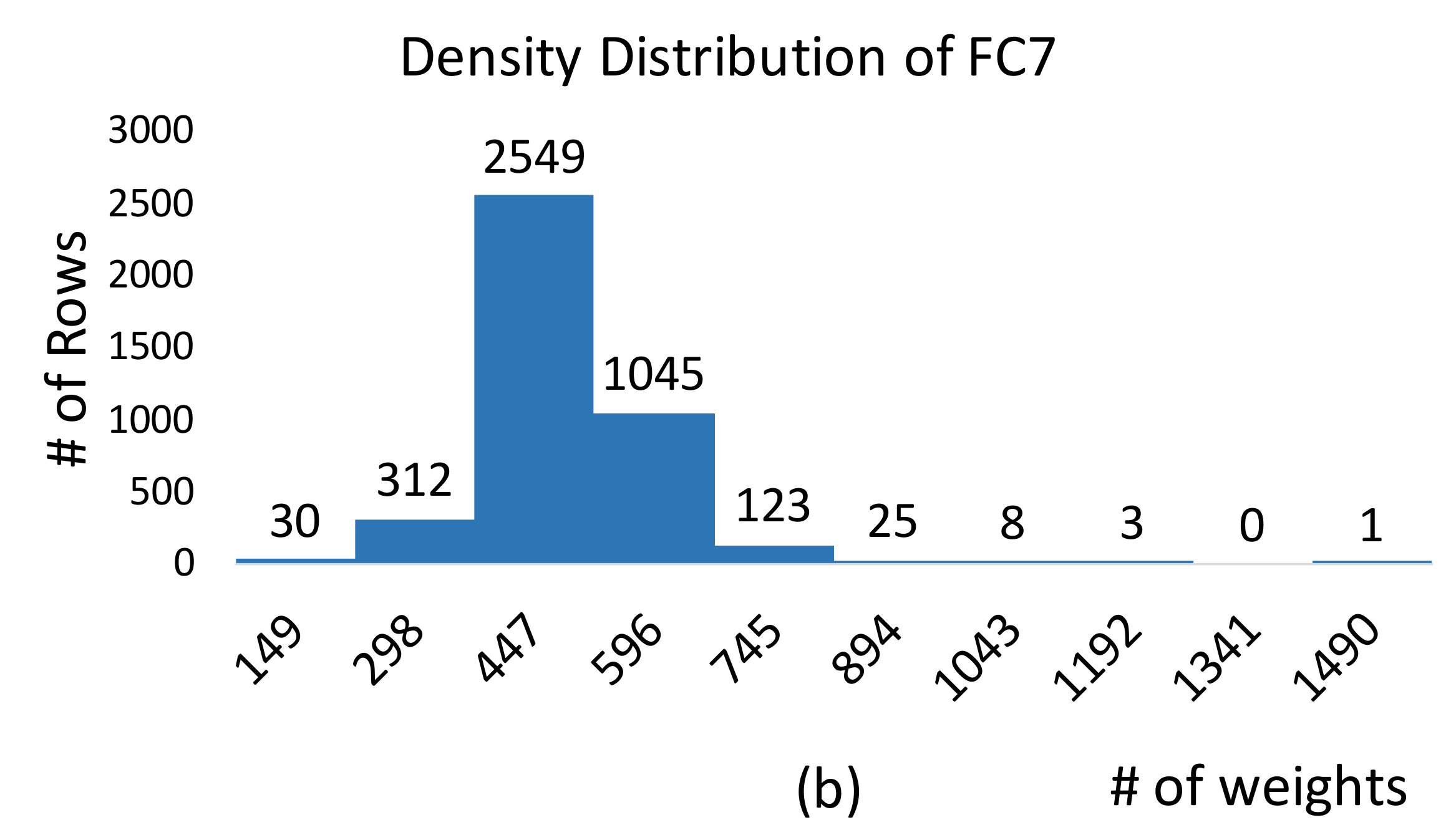}
\endminipage
\minipage{0.313\textwidth}%
  \includegraphics[width=\linewidth]{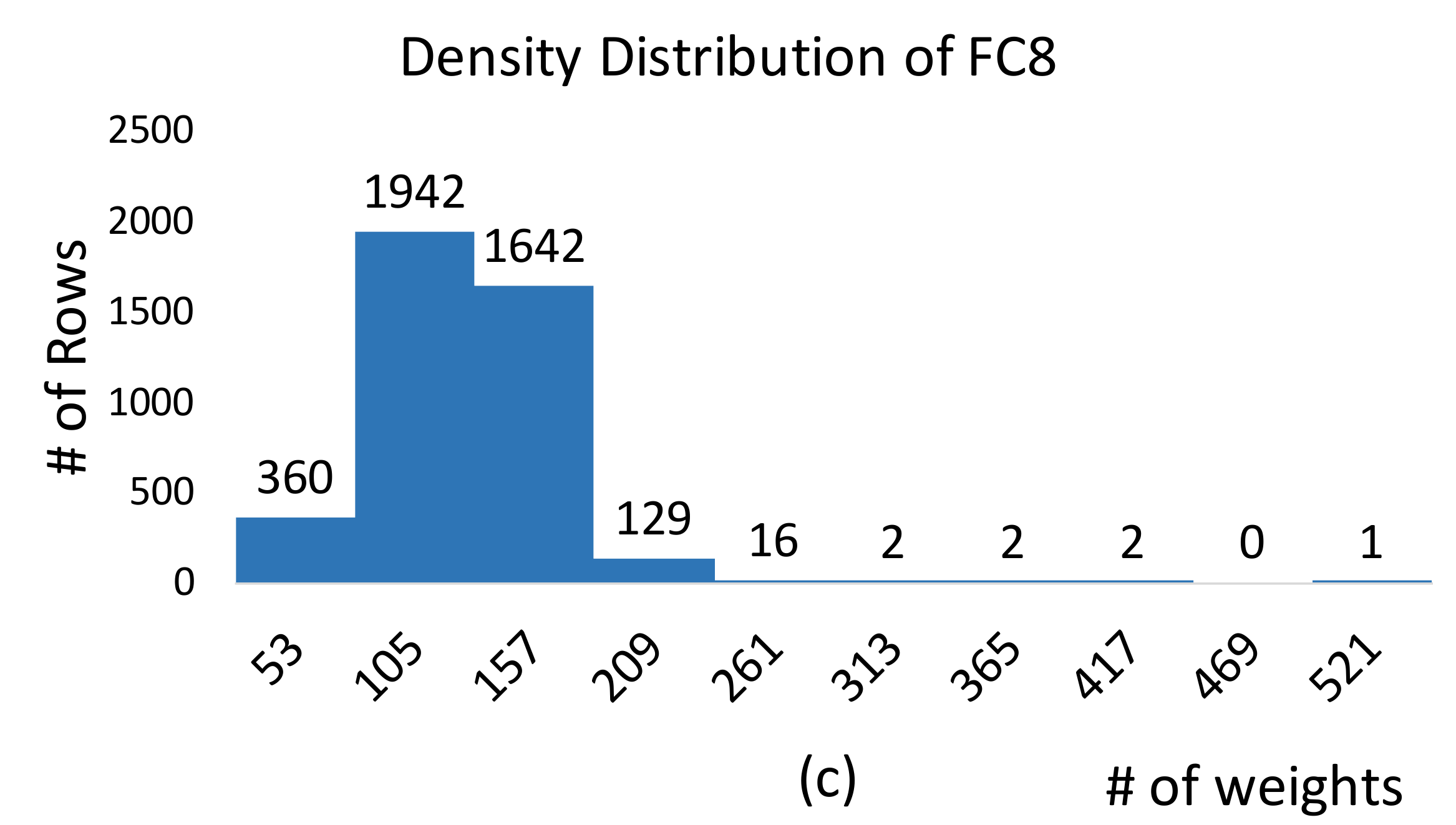}
\endminipage
\caption{Row density distribution of FC6-FC8 of AlexNet with 90\% sparsity: (a) FC6, 
(b) FC7, (c) FC8.}\label{fig:alexnet_90}
\end{figure*}

Structured pruning techniques are proposed to address the drawbacks of irregular pruning, and they
can be categorized into two types.

The first type of structured pruning techniques requires no index or decoding, such as matrix
factorization-based methods \cite{thakker2019compressing,sainath2013low,kim2019efficient},
block-circulant algorithm \cite{wang2018c}, and PermDNN \cite{deng2018permdnn}. The low-rank matrix
factorization proposed in \cite{sainath2013low} factorizes the original high-rank weight matrix into
low-rank sub-matrices and thereby achieves 30\%-50\% parameter reduction.
\cite{thakker2019compressing} proposes to use Kronecker Product to decompose the high-rank matrix.
However, this method achieves only little computation reduction, because the Kronecker Product
computation to reconstruct a high-rank matrix from the decomposed sub-matrices is not trivial.  

Block-circulant pruning proposed in \cite{wang2018c} divides the original weight
matrix into sub-blocks and enforces each block to be a circulant matrix. Since this method requires
storing only the first row of each circulant block, it can reduce the space complexity from $O(n^2)$
to $O(n)$ and the computation complexity from $O(n^2)$ to $O(n\log n)$. However, the computation is
mainly performed in the frequency domain, which significantly diminishes the benefit of computation
reduction because of the Fast Fourier Transformation and complex domain computation. PermDNN
\cite{deng2018permdnn} decomposes the weight matrix into blocks and each block is a permuted
diagonal matrix. Nevertheless, this method considers only the simplification of hardware
implementation, resulting in limited pruning ratio.

The other type of structured pruning works in the way of sharing index among multiple neighboring
weights, so that the index storage is reduced and the {\em decoding efficiency}\footnote{Decoding
efficiency is defined as the number of activations selected per clock cycle for the corresponding
retained weights.} can be improved.
Column pruning removes one column of the weight matrix \cite{wen2017learning}. Although this method
is effective in convolutional layers, it fails to work in the fully-connected layers, where removing
one column can cause significant information loss as it is equivalent to removing one input
activation. Prior work \cite{wen2017learning,wang2019acceleration} adopts this strategy in RNNs but
only achieves about 2$\times$ parameter reduction. Block pruning performs pruning at the scale of
blocks \cite{van2019rethinking}, but grouping neighboring weights into a specific structure is a
strong constraint which is not an effective way to keep the salient weights. As a result, only
limited pruning ratio can be achieved. 



\section{Methodology}

Based on the survey of the related work, we can conclude that (i) irregular pruning can achieve the
highest compression ratio among all compression techniques mentioned above; (ii) structured pruning
works in the way of trading off the regularity with pruning ratio at a given accuracy. In order to
make the pruned weight matrix structural, two types of constraints are usually applied. One is
to enforce all rows in the weight matrix to have the same number of weights retained, and the other
is to group neighboring weights into certain structures, such as the block pruning
\cite{van2019rethinking} or block-circulant-based compression \cite{wang2018c}.

Even though they can more or less achieve trade-off improvement, those previous studies have never
discussed the fundamental question: \emph{whether or not the the hardware-friendly algorithm can lead to 
high pruning ratio, which is another vital factor for achieving high performance and energy efficiency.} To answer this
question, it is favorable to explore the intrinsic characteristics of the pruned neural networks to
help us understand the essence of pruning. Since irregular pruning can achieve the highest
compression ratio, in this work we study the structure characteristics of irregularly pruned weight
matrices. In addition, we ultimately propose a structured pruning method that can achieve both high
pruning ratio and decoding efficiency without accuracy loss.

\subsection{Structure Characteristics of a Weight Matrix}

In this work, we first obtain the pruning mask by applying the irregular pruning into the matrix.
The pruning mask is a bitmap where `0' (`1') indicates if the weight needs to be pruned
(retained). We then examine the row density distribution and the position distribution of the
weights within each row. For the study, two models are selected and each is the representative of a
specific domain. The first model is a medium size LSTM for language
modeling~\cite{zaremba2014recurrent}, which has the architecture of (embedding:
10k$\times$650)--(LSTM: 650)--(LSTM: 650)--(decoder: 650$\times$10k). The second model is AlexNet
for image classifications~\cite{Krizhevsky2012}, and all of its fully-connected layers (i.e.,
FC6-FC8) are studied.

Figure~\ref{fig:lstm_90} and~\ref{fig:alexnet_90} plots the row density distribution of the LSTM
model and AlexNet (FC6-FC8 only), respectively. Both neural networks have been pruned irregularly to
reach 90\% sparsity. From the two figures, we can find that the row density distributions vary
significantly not only across different neural networks, but also across different layers within the
same network, and even across different rows within the same layer. For example, the densest rows
can have up to 14$\times$ more weights than the sparsest rows.

Nowadays, enforcing all rows in a layer to have the same density is widely adopted to assure the
regularity. Nevertheless, due to our observation on the wide variation in the row density
distribution, we question this as a reasonable strategy. Therefore, we further investigate the
sensitivity of a row for further pruning.
We have done the experiments by firstly performing irregular pruning with an overall 70\% pruning
ratio to obtain the pruning mask. The density of each row is then calculated (i.e., counting 1s in
the mask) and sorted.  We evenly divide the sorted rows into two segments, the first half is denoted
as \emph{sparse rows} and the second half as \emph{dense rows}. Then, we anchor the pruning ratio of
the dense rows and increase the pruning ratio of the sparse rows. The model is then retrained to
recover the perplexity\footnote{Perplexity is the metric to evaluate a language model and it is the lower the better.}. 
As a comparison, we also anchor the pruning ratio of the sparse rows while
increasing the ratio of the dense rows. Please refer to Section~\ref{sec:evaluation} for the detail
of experiment setup. 

\begin{figure}[t]
\centering
\includegraphics[width=0.42\textwidth]{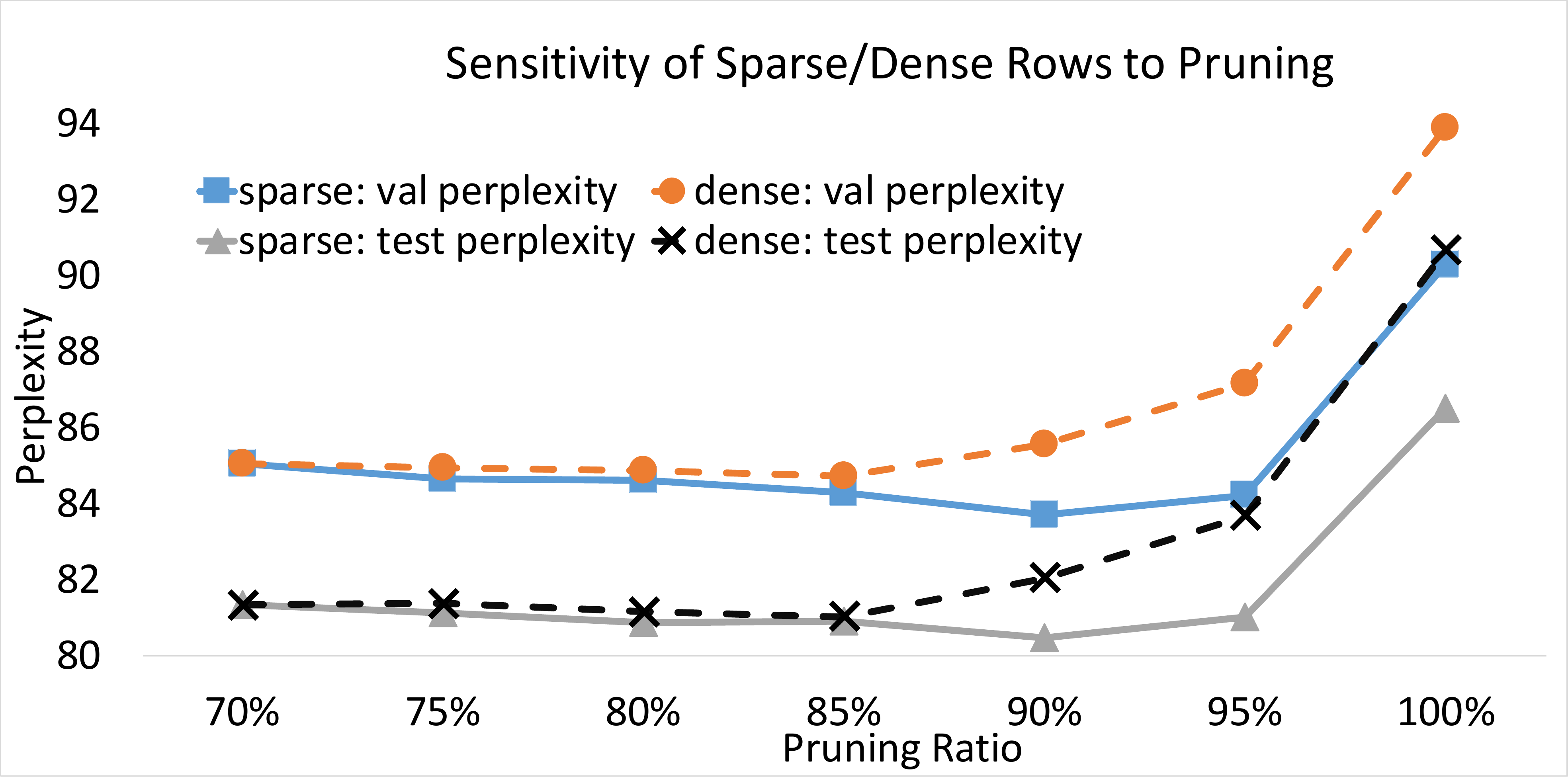}
\caption{Sensitivity study of sparse and dense rows. The dense rows are more sensitive to further
pruning as the solid curves are always above the dashed curves.}
\label{fig:pruning_sensitivity}
\end{figure}

Figure~\ref{fig:pruning_sensitivity} illustrates the sensitivity of the sparse and dense rows to
pruning ratio, respectively. The dashed curves show the results when sweeping pruning ratio in the
dense rows. The solid curves show the results when sweeping pruning ratio in the sparse rows. The
validation and test perplexities are plot separately. As shown, the dense rows are more sensitive to
further pruning than the sparse rows, as the solid curves are always above the dashed curves.
Generally, the sparse rows are resilient to further pruning. Thanks to the regularization effect,
the model performance can be even improved with moderate pruning ratio (e.g., 70\%-90\%). However,
as the pruning ratio goes high enough (e.g., $\geq$95\%), many rows are pruned entirely, which leads
to a sharp performance degradation.

In addition to the distribution of row density, we also conduct the characterization study on weight
positions. In the study, we first extract the pruning mask by performing irregular pruning with 90\%
pruning ratio on the medium size LSTM. Note that only the pruning mask is generated at this step, while the matrix
is not really pruned. According to the pruning mask, we then measure the relative distance between two
adjacent non-zero weights in the same row. We figure out that the average relative distance in all
layers is ten, which corresponds to the sparsity (i.e., one out of ten weights is retained for 90\%
pruning ratio).  As summarized in Table~\ref{Table:hit_rate}, when we divide each row into blocks
whose sizes are ten, about 26\% blocks across all layers have more than one non-zero weights, around
39\% blocks have exactly one non-zero weight, and the remaining 35\% blocks are empty.

To determine if those empty blocks contain only in-salient weights, we define a new metric
\emph{relative difference} to quantify the significance of weights in the empty blocks. Firstly, we
pick the largest weight in each empty block and calculate the arithmetic mean of their absolute
values as $\bar{W}_{empty}$. We then turn to the blocks that have more than one weights and pick all
retained weights but the largest one (all-but-largest, or abl for short) to calculate their mean
value as $\bar{W}_{abl}$. Finally, we compare the two mean values and denote the value
$\frac{|\bar{W}_{abl}-\bar{W}_{empty}|}{\bar{W}_{abl}}$ as their relative difference. In this way, a
large relative difference indicates that empty blocks only have in-salient weights while a small
difference implies that empty blocks also have salient weights, even though they may be slightly
less important than the retained weights selected by irregular pruning.

\begin{table}[t]
\centering
\caption{The portion of blocks that contain different number of weights. Block size is ten.}\label{Table:hit_rate}
\resizebox{0.98\columnwidth}{!}{
\begin{tabular}{ccccc}
\hline
Layer  & \makecell{\# of weights\\=0}  & \makecell{\# of weights\\=1} & \makecell{\# of weights\\$>$ 1} \\                
\hline
Embedding	& 34.86\%	& 38.79\%	& 26.35\%              \\
\hline
LSTM1		& 34.85\%	& 38.75\%	& 26.40\%              \\
\hline
LSTM2		& 34.94\%	& 38.63\%	& 26.43\%              \\
\hline
Decoder		& 34.95\%	& 38.66\%	& 26.39\%             \\
\hline
\end{tabular}
}
\end{table}


In our experiment, the relative difference is about 27.6\% for both embedding and decoder layers,
and 12.3\% for the LSTM layers. As the relative difference is small, it reveals that these empty
blocks also have salient weights. This finding inspires us to reform the weight selection strategy
by devising the new \emph{block-max weight masking (\bmwm)} method. In contrast to the irregular
pruning, \bmwm~simply picks the largest weight in magnitude from each block. Specifically, 
\bmwm~consists of three steps: (1) It first divides a row into equally sized blocks. For instance, given a
row that has 1000 elements and the block size is 10, the row should have 1000/10=100 blocks. (2)
Within each block, it retains the largest weight in magnitude. (3) It repeats (1) and (2) for each
row of the weight matrix.

To demonstrate the effectiveness of \bmwm, we compare the salience of the retained weights after
\bmwm~with that after irregular pruning. Specifically, the mean value of the magnitude of the
retained weights after \bmwm~is calculated as $\bar{W}_{bmwm}$, and that after irregular pruning is
$\bar{W}_{irr}$, their relative difference is calculated as
$\frac{|\bar{W}_{irr}-\bar{W}_{bmwm}|}{\bar{W}_{irr}}$.
As shown in Table~\ref{Table:pruning_difference}, the relative difference for the embedding layer
and decoder layer is reduced to 8.8\%, and that for the LSTM layers drops to 4.3\%. In other words,
\bmwm~can effectively sustain the salience of the weight matrix, similar to irregular pruning. On
the other hand, we also evaluate the salience of the retained weights selected by the
state-of-the-art block pruning method with 4$\times$4 block size.  The relative difference between
block pruning and irregular pruning is 45\% in the embedding and decoders layer, and that in the
LSTM layers is 31\%, which is much larger than \bmwm. Therefore, we are confident that \bmwm~is
simple yet effective in selecting the salient weights.


\begin{figure}[tbp]
\centering
\includegraphics[width=0.42\textwidth]{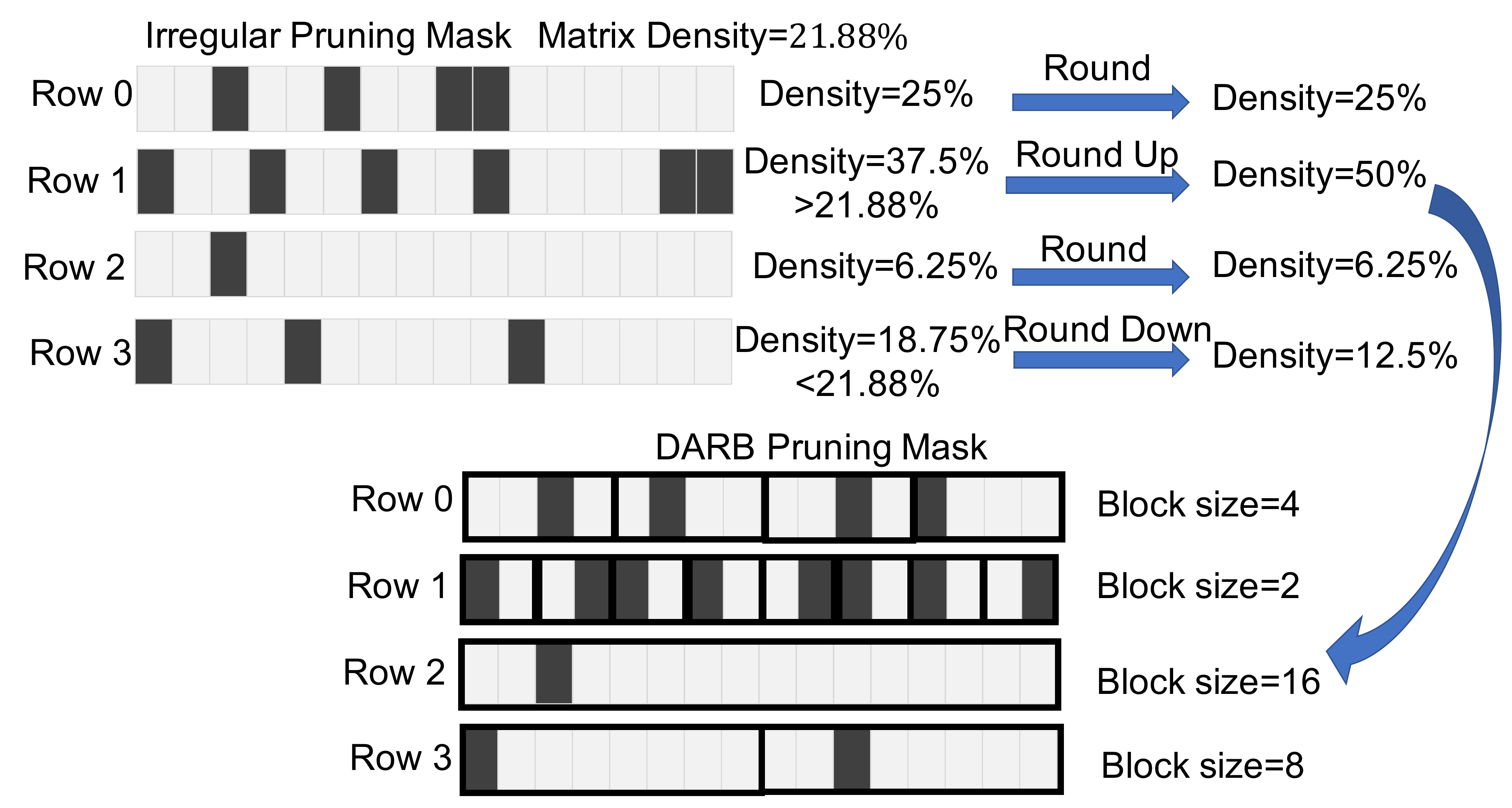}
\caption{The generation of a \darb~pruning mask. Irregular pruning mask is used as the starting
point.} \label{fig:darb_pruning}
\end{figure}

\begin{table}[t]
\centering
\caption{The Comparison of Relative Difference of Retained Weights between \bmwm~and Block Pruning}\label{Table:pruning_difference}
\resizebox{\columnwidth}{!}{
\begin{tabular}{ccc}
\hline
Layer & \makecell{\bmwm\\to Irregular Pruning} & \makecell{Block Pruning $4\times4$\\to Irregular Pruning}   \\                
\hline
Embedding	& 8.8\%	& 45.3\%	              \\
\hline
LSTM1		& 4.3\%	& 31.1\%	              \\
\hline
LSTM2		& 4.3\%	& 31.1\%	              \\
\hline
Decoder		& 8.9\%	& 45.2\%	             \\
\hline
\end{tabular}
}
\end{table}

\subsection{Density-Adaptive Regular-Block Pruning}


As discussed above, maintaining the variety of row density helps sustain accuracy.
Therefore, it is desirable to consider this characteristic when designing a structured
pruning method.
However, maintaining this variety incurs difficulty in storing the retained weights in a
compact and regular manner. In addition, even though \bmwm~is effective to select salient weights,
the block size can be hardware unfriendly. To fully exploit the discovered characteristics of neural
networks, we propose a density-adaptive regular-block (\darb) pruning method.

As illustrated in Figure~\ref{fig:darb_pruning}, \darb~first needs to obtain the pruning mask
generated in irregular pruning. Note that this step is to calculate the row density, rather than
perform real pruning. Like \bmwm, \darb~then divides each row into blocks and keeps only the weight
that has the largest magnitude in each block. Distinct from \bmwm, \darb~allows different rows to
adopt different block sizes so that the irregularity of row density can be maintained to some
extent. To make it hardware friendly, \darb~intentionally constrains the block size to be in power
of two, such as 2, 4, ..., etc.

To obey the constraint on block size, the row density sometimes needs to be adjusted. Given that
dense rows are more sensitive to further pruning, maintaining the density of dense rows helps
sustain the model accuracy better. Also, we need to avoid the density from being rounded in the same
direction that can change the pruning ratio significantly. Therefore, when the density of a row is
greater (less) than the matrix density, it is rounded up (down). For instance, suppose the matrix
density after irregular pruning is 21.88\%, if the density of a row is 37.5\%, it is rounded up to
50\% so that the block size is two (=1/0.5). On the other hand, if the density of a row is 18.75\%,
it needs to be rounded down to 12.5\% and the resulting block size is eight (=1/0.125).

\begin{figure}[tbp]
\centering
\includegraphics[width=0.42\textwidth]{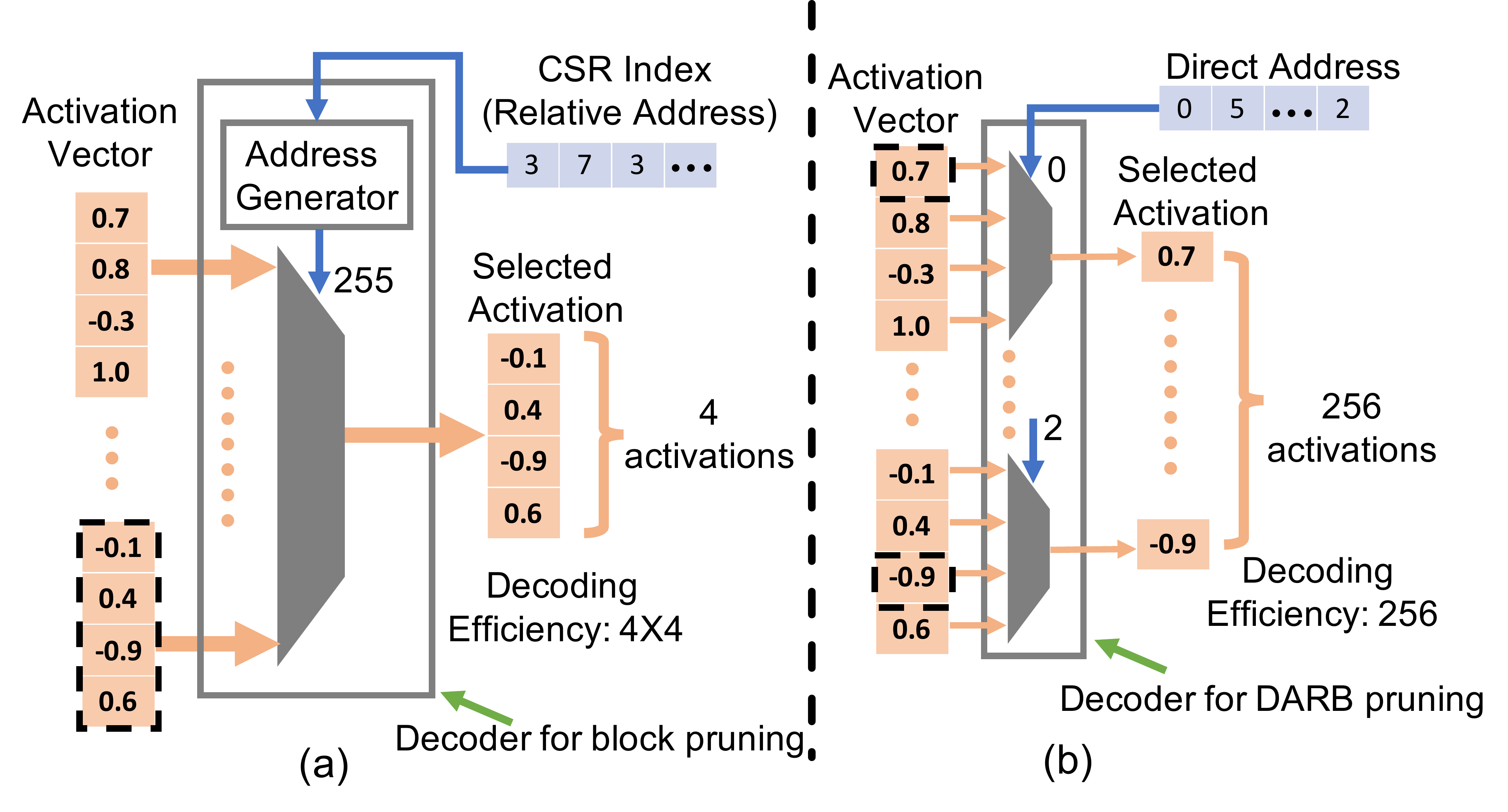}
\caption{Decoding efficiency for (a) block pruning, (b) \darb~pruning.} \label{fig:decoding_efficiency}
\end{figure}

In this way, \darb~can achieve excellent trade-off between regularity and pruning ratio with the
following benefits. Firstly, the block size is hardware friendly to facilitate index decoding. As
illustrated in Figure~\ref{fig:decoding_efficiency}(a), the conventional CSR encoded indices need a
big multiplexer to accomplish the indexing, which makes the entire decoding process quite slow since
it can only be done in sequence. Even though block pruning can share the index among multiple
weights, it is still hard to significantly improve the decoding performance since the block size is
substantially limited by the consideration of accuracy. In contrast, thanks to the small
equally-sized blocks, \darb~supports highly efficient decoding by leveraging multiple small
multiplexers in parallel, as shown in Figure~\ref{fig:decoding_efficiency}(b). Suppose the row size
 is 1,024, and the block size is four, \darb~can simply select 256 activations
in parallel with similar area cost.



Secondly, 
the constraint on the block size enforces the row density to be clustered into different groups, and
the density ratio between two groups is still in the power of two. Figure~\ref{fig:darb_pruning}
explains the motivation. In the figure, the density of row 0 and 1 is 25\% and 50\%, respectively.
Therefore, the density of row 1 is $2^1$ times of row 0. Such regularity among row densities 
eventually facilitates the weights storing. 

\section{Evaluation} \label{sec:evaluation}

To evaluate the efficacy of the proposed \darb~pruning, we test it on five neural networks. The
pruning ratio and the accuracy of the pruned models are compared with the state-of-the-art work,
including both irregular pruning and structured pruning methods. In addition, the efficiency of
index decoding is also evaluated by counting the number of activations that can be selected per
clock cycle in an exemplary hardware implementation.
In this work, we adopt the ADMM-based pruning framework \cite{ren2019admm}, which has achieved the
state-of-the-art irregular pruning ratio. The ADMM-based pruning framework decomposes the pruning
problem into two subproblems, one is to find a good pruning mask, and the other is to train the model
via this mask. By iteratively solving these two subproblems, the target pruning ratio can be
achieved. 

\subsection{Algorithm Evaluation}
\subsubsection{Language Modeling.} 
For this task, a large size two-layer LSTM \cite{zaremba2014recurrent} is built to perform the
word-level prediction for Penn Tree Bank (PTB) dataset~\cite{marcus1993building}, whose vocabulary
size is 10k words. The training data, validation data, and test data of PTB dataset has 929k, 73k,
and 82k words, respectively. The architecture of the model is (embedding: 10k$\times$1500)--(LSTM:
1500)--(LSTM: 1500)--(decoder: 1500$\times$10k). The model is trained with 20 batches and 35
unrolling steps, which has the same configurations as the prior arts
\cite{zaremba2014recurrent,wen2017learning}. The dropout configuration for \darb~is (0.35, 0.75) in
the ADMM regularization step and (0.35, 0.7) in the retrain step, where the former in the parentheses is the dropout
for LSTM layers, and the latter is for other layers. For the sake of fairness, two pruning ratios
are applied separately to make sure that each achieves the similar perplexity to its counterpart. 

\begin{table}[t]
\centering
\caption{Pruning Results and Comparison on Large Size LSTM for PTB}\label{Table:lstm1500}
\resizebox{0.97\columnwidth}{!}{
\begin{tabular}{ccccc}
\hline
Method 	&  Perplexity       & Para. No. & \makecell{Pruning\\Ratio}  \\ 
\hline
baseline  &  (82.20, 78.40)   &  66.00M	&  1.00$\times$  \\
\hline
Irregular	&  (82.20, 78.40)   &  3.00M   		&  20.00$\times$  \\
\hline
ISS	&  (82.59, 78.65)   &  21.80M   		&  3.03$\times$  \\
\hline
BBS  & (---, 79.20) 	   & ---  		&  4.00$\times$ \\
\hline
\bf{\darb-1 (ours)} & (82.15, 78.51) & 5.02M  & 13.14$\times$   \\
\hline
\bf{\darb-2 (ours)} & (82.65, 79.17) & 4.26M  & 15.48$\times$   \\
\hline
\end{tabular}
}
\end{table}

Table \ref{Table:lstm1500} shows the comparison between \darb~and the state-of-the-art pruning
methods on the LSTM model. The first number in the perplexity tuple is the validation perplexity,
and the second is the test perplexity. Compared with ISS~\cite{wen2017learning} , which has similar
perplexity with \darb-1, \darb-1 achieves 13.14$\times$ pruning ratio and outperforms ISS by
4.34$\times$. In addition, \darb-2 can achieve 15.48$\times$ pruning ratio, which is 3.78$\times$
higher than BBS~\cite{cao2019efficient}. Note that ISS can only prune the LSTM and decoder layers
but leave the embedding layer untouched due to the high sensitivity to accuracy loss\footnote{It is
unclear if BBS can prune the entire network.}. Instead, \darb~can prune all four layers of the model
by 13.14$\times$ and meanwhile sustain the model accuracy. Despite the 20$\times$ pruning ratio, the
extremely low decoding efficiency makes irregular pruning inapplicable in real applications. We will
discuss the decoding efficiency at the end of this section.

\subsubsection{Speech Recognition.}
This task is evaluated with TIMIT \cite{garofolo1990timit}, which is an acoustic-phonetic speech
corpus. It contains broadband recordings from 630 speakers, and each speaker reads ten phonetically
rich sentences in eight major dialects of American English. We build two models for the task. The
first model is a projected LSTM (LSTMP)~\cite{zia2019long}, which has two LSTM layers with 1,024
hidden units and a projection layer with 512 hidden units. The second model is a two-layer gated
recurrent unit (GRU) with 1,024 hidden units in each layer. For this task, the evaluation metric is
phone error rate (PER), which is the smaller the better.

\begin{table}[t]
\centering
\caption{Pruning Results ad Comparison on LSTMP for TIMIT}\label{Table:lstmp}
\resizebox{\columnwidth}{!}{
\begin{tabular}{ccccc}
\hline
Method 	&  \makecell{PER\\Degradation}       & Para. No. &
\makecell{Pruning\\Ratio}  \\ 
\hline
baseline   &  20.70\% $\to$ 20.70\%   &  3.25M	&  1.0$\times$  \\
\hline
Irregular   &  20.70\% $\to$ 20.90\%   &  0.16M	&  20.0$\times$  \\
\hline
C-LSTM   & 24.15\% $\to$ 24.57\%   &  0.41M  		&  8.0$\times$  \\
\hline
C-LSTM   & 24.15\% $\to$ 25.48\%   &  0.20M  		&  16.0$\times$  \\
\hline
BBS    & 23.50\% $\to$ 23.75\%  &  0.41M  		&  8.0$\times$ \\
\hline
\bf{\darb-a (ours)} &  20.70\% $\to$ 20.80\% &  0.41M  &  8.0$\times$  \\
\hline
\bf{\darb-b (ours)} &  20.70\% $\to$ 20.90\% &  0.20M  & 16.0$\times$   \\
\hline
\bf{\darb-c (ours)} &  20.70\% $\to$ 21.00\% &  0.16M  &  20.0$\times$  \\
\hline
\end{tabular}
}
\end{table}

\begin{table}[t]
\centering
\caption{Pruning Results and Comparison on GRU for TIMIT}\label{Table:gru}
\resizebox{0.98\columnwidth}{!}{
\begin{tabular}{ccccc}
\hline
Method 	  &  PER       & Para. No. & Pruning Ratio  \\ 
\hline
baseline  &  18.8\%   &  12.7M	&  1.0$\times$  \\
\hline
Irregular  &  19.0\%   &  0.40M	&  32.0$\times$  \\
\hline
Block-circulant-1	&  19.4\%   & 1.59M   &  8.0$\times$  \\
\hline
Block-circulant-2	&  20.1\%   & 0.79M   &  16.0$\times$  \\
\hline
\bf{\darb-A} & 18.8\% & 1.59M  &  8.0$\times$  \\
\hline
\bf{\darb-B} & 19.1\% & 0.40M  &  32.0$\times$  \\
\hline
\end{tabular}
}
\end{table}

Table~\ref{Table:lstmp} shows the comparison results. Note that \darb~has three variants with
different pruning ratios. For this model, we prune only one layer of LSTMP, same as the previous
work. In addition, since those prior studies have different baselines, we present PER degradation
after pruning. When pruning ratio is 8$\times$, the best prior work is BBS, which has 0.25\% PER
degradation. \darb-a instead has only 0.1\% degradation. Comparing to C-LSTM~\cite{wang2018c} that
achieves 16$\times$ pruning ratio with 1.32\% PER degradation, \darb-b can achieve the same pruning
ratio with only 0.2\% degradation. When the pruning ratio increases to 20$\times$, \darb~has only
0.3\% PER degradation, which is 0.1\% worse than irregular pruning. As a result, the performance of
\darb~is close to irregular pruning.

Table~\ref{Table:gru} shows the pruning results on the GRU model, and \darb~is compared with the
block-circulant algorithm. The Block-circulant algorithm sees 0.6\% and 1.3\% PER degradation with
8$\times$ and 16$\times$ pruning ratio, respectively. In contrast, \darb-A (8$\times$ pruning ratio)
has no degradation and \darb-B (32$\times$ pruning ratio) has only 0.3\% degradation. Furthermore,
\darb-B even beats the block-circulant algorithm with 8$\times$ pruning ratio. Again, \darb~achieves
similar PER performance as irregular pruning.

\subsubsection{Image Classification.}
To evaluate the feasibility of \darb~on non-NLP models, we train an AlexNet \cite{Krizhevsky2012}
with ImageNet \cite{deng2009imagenet}. Table~\ref{Table:alexnetfc} gives the comparison between
\darb~and PermDNN~\cite{deng2018permdnn} when pruning FC6-FC8 of AlexNet.  Note that the baseline of
PermDNN is the vanilla AlexNet, while ours is AlexNet-BN that is trained with batch-normalization.
Although the baselines are different, the original AlexNet-BN has higher accuracy, which is
intuitively more difficult to prune. To assure fair comparison, we focus on the relative accuracy
loss.
As shown, \darb-I achieves 21.3$\times$ pruning ratio with no accuracy loss, which is remarkable
because it is close to irregular pruning (24$\times$), while PermDNN has 0.2\% accuracy loss with
9$\times$ pruning ratio. If the same accuracy loss is allowed, \darb-II is able to achieve
25$\times$ pruning.

Also, we evaluate \darb~in the convolutional layers of VGG-16 on CIFAR-10. $18.7\times$ pruning
ratio is obtained with 0.6\% accuracy degradation. The result suggests that \darb~is also promising
for CNNs. We will optimize \darb~for state-of-the-art CNNs as our future work.

\begin{table}[t]
\centering
\caption{Pruning Results and Comparison on FC6,FC7,FC8 of AlexNet for ImageNet}\label{Table:alexnetfc}
\resizebox{0.96\columnwidth}{!}{
\begin{tabular}{cccc}
\hline
Method 		  &  \makecell{Top-5 Acc.}   & Para. No. & \makecell{Pruning
Ratio}  \\ 
\hline
AlexNet     &  80.2\%   &  58.6M	&  1.0$\times$  \\
\hline
\makecell{PermDNN on\\AlexNet}	  &  80.0\%   &  6.5M  &  9.0$\times$  \\
\hline
\hline
AlexNet-BN     &  83.4\%   &  58.60M	&  1.0$\times$  \\
\hline
Irregular     &  83.4\%   &  2.44M	&  24.0$\times$  \\
\hline
\makecell{\bf{\darb-I on}\\\bf{AlexNet-BN}}    & 83.4\% & 2.75M  &  21.3$\times$  \\
\hline
\makecell{\bf{\darb-II on}\\\bf{AlexNet-BN}}    & 83.2\% & 2.34M  &  25.0$\times$  \\
\hline
\end{tabular}
}
\end{table}

\subsection{Demonstration of Efficient Index Decoding}\label{decoding_results}


To demonstrate the index decoding efficiency of \darb, we define the decoding efficiency as the
number of activations that can be selected from an activation vector per clock cycle. Block pruning
is used as our baseline. The size of the activation vector in the aforementioned LSTM model is 1500.
We implement six decoders in RTL to concurrently support the block size of 2, 4, ..., and 64. They
are synthesized in CMOS 40$nm$ process to measure the hardware area. We also design and synthesize
the decoders for the baseline to support two types of block pruning. As shown in
Table~\ref{Table:decoding_efficiency}, given the same hardware area, we can equivalently deploy 6.41
and 6.45 decoders for the block-pruning design whose block size is 4$\times$4 and 8$\times$8,
respectively. From the table, we can find that the decoding efficiency of \darb~outperforms the
other two designs by up to 14.3$\times$. Moreover, once no perplexity degradation is allowed,
\darb~can achieve 1.76$\times$ and 2.45$\times$ higher pruning ratio than the other two designs.  It
is worth noting that the decoding efficiency of irregular pruning is only 7 with the same footprint.
Such low decoding efficiency explains why it is rarely adopted in reality.

Regarding the index storage overhead, block pruning only has minor advantage over \darb. In
general, block pruning has significantly lower pruning ratio than \darb, which indicates that its
overall storage is still dominated by the weight itself. For instance, the original model has 66
million weights. After block pruning, 12.29 million weights (v.s. \darb's 5.02 million) are retained so that 0.2(=12.29/64)
million indices are required. 
Assuming weights are quantized into INT8, it needs at least 12.29MB to
store weights. Therefore, even if it only needs $<$0.1MB for relative CSR index, the overall storage saving is minor. 
On the other hand, \darb~needs 5.02(=66/13.14) million indices. It seems huge at
the first glance. However, the weight index in \darb~just indicates its position within
the block. As the block size is a power of two, a block whose size is $m$ only requires $\log_2
m$ bits for indexing. Therefore, the average bit width of each index is less than four, and the
total index storage overhead of \darb~is less than 0.63MB, which is slightly larger than block
pruning but still minimal. 

\begin{table}[t]
\centering
\caption{Decoding Efficiency and Pruning Ratio Comparison Between \darb~and Block Pruning}\label{Table:decoding_efficiency}
\resizebox{0.98\columnwidth}{!}{
\begin{tabular}{cccc}
\hline
Metric 	  &  \darb   & Block-4$\times$4  & Block-8$\times$8   \\ 
\hline
\makecell{Area (${\mu}m^2$)}  &  58928   &  58928	& 58928   \\
\hline
\makecell{Decoder No.}	      & 6    & 6.41    &  6.45   \\
\hline
\makecell{Decoding Efficiency\\ (activations/cycle)} 	&  1478   & 103    & 413    \\
\hline
\makecell{Pruning Ratio}    & 13.14$\times$    & 7.48$\times$     & 5.37$\times$ \\
\hline
\end{tabular}
}
\end{table}

\section{Conclusion}

In this work, we study the intrinsic characteristics of irregularly pruned weight matrices,
including the row density distribution, the sensitivity of different rows to further pruning, and
the positional characteristic of weights within each row. Motivated by our observations, we propose a
block-max weight masking method that is superior in selecting salient weights than the state-of-the-art prior. 
Then we devise a density-adaptive regular-block pruning method that can simultaneously achieve high pruning ratio and decoding efficiency, with almost no accuracy loss. Experimental results show that \darb~outperforms
the state-of-the-art prior in terms of pruning ratio by 2.8$\times$ to 4.3$\times$.
And it achieves up to 14.3$\times$ decoding efficiency over the block pruning.

On the other hand, inspired by prior work \cite{liu2018rethinking,frankle2018lottery}, we believe
that mining more characteristics of sparse neural networks has great potential in not only pushing the state-of-the-art pruning ratio,
but also facilitating neural architecture search. We will extend our research in this direction.

\fontsize{9.0pt}{10.0pt} \selectfont
\bibliographystyle{aaai}
\bibliography{references}

\end{document}